\documentclass[lettersize,journal]{IEEEtran}
\usepackage{amsmath,amsfonts}
\usepackage{algorithmic}
\usepackage{algorithm}
\usepackage{array}
\usepackage[caption=false,font=normalsize,labelfont=sf,textfont=sf]{subfig}
\usepackage{textcomp}
\usepackage{stfloats}
\usepackage{url}
\usepackage{verbatim}
\usepackage{graphicx}
\usepackage{cite}
\usepackage{color}
\usepackage{xcolor}
\usepackage{bm}
\usepackage{booktabs}
\usepackage{multirow}
\begin{document}

\title{Identity Clue Refinement and Enhancement for Visible-Infrared Person Re-Identification}

\author{Guoqing Zhang,
        Zhun Wang, 
        Hairui Wang,
        Zhonglin Ye,
        and
        Yuhui Zheng
        

\thanks{This research was supported by the National Natural Science Foundation of China under Grant 62172231, 92470202 and U22B2056; and by the Natural Science Foundation of Jiangsu Province of China under Grant BK20220107.}
        
\thanks{Guoqing Zhang is with the School of Computer Science, Nanjing University of Information Science and Technology, and also with the Key Laboratory of Social Computing and Cognitive Intelligence (Dalian University of Technology), Ministry of Education, Dalian, China (e-mail: guoqingzhang@nuist.edu.cn).}
\thanks{Zhun Wang and Hairui Wang are with the School of Computer Science, Nanjing University of Information Science and Technology, Nanjing, 210044, China (e-mail: wangzhun0530@outlook.com; hairuiwang@nuist.edu.cn).}

\thanks{Yuhui Zheng and Zhonglin Ye are with the State Key Laboratory of Tibetan Intelligent Information Processing and Application, Qinghai Normal University, Xining, 810008, China (e-mail: zhengyh@vip.126.com; yezhonglin@qhnu.edu.cn). (Corresponding Author: Zhonglin Ye)}

}

\maketitle

\begin{abstract}
Visible-Infrared Person Re-Identification (VI-ReID) is a challenging cross-modal matching task due to significant modality discrepancies. While current methods mainly focus on learning modality-invariant features through unified  embedding spaces, they often focus solely on the common discriminative semantics across modalities while disregarding the critical role of modality-specific identity-aware knowledge in discriminative feature learning. To bridge this gap, we propose a novel Identity Clue Refinement and Enhancement (ICRE) network to mine and utilize the implicit discriminative knowledge inherent in modality-specific attributes. Initially, we design a Multi-Perception Feature Refinement (MPFR) module that aggregates shallow features from shared branches, aiming to capture modality-specific attributes that are easily overlooked. Then, we propose a Semantic Distillation Cascade Enhancement (SDCE) module, which distills identity-aware knowledge from the aggregated shallow features and guide the learning of modality-invariant features. Finally, an Identity Clues Guided (ICG) Loss is proposed to alleviate the modality discrepancies within the enhanced features and promote the learning of a diverse representation space. Extensive experiments across multiple public datasets clearly show that our proposed ICRE outperforms existing SOTA methods.
\end{abstract}


\section{Introduction}
\IEEEPARstart{P}{erson} Re-identification (ReID) aims to match the same individual across diverse temporal and spatial contexts by analyzing images captured by cameras with non-intersecting views \cite{r42, r45}. With the growing public concern for societal security, as an important part of the monitoring systems, it has attracted considerable interest from scholars and made remarkable progress. However, most existing methods solely choose RGB images as the carrier for pedestrian description, neglecting the interference of low-light environments on the imaging quality of RGB images, so the application scenarios are limited and primarily effective under bright daylight conditions. To achieve round-the-clock ReID, recent works have integrated clear pedestrian images captured by infrared cameras in dimly lit settings, proposing an important yet challenging task (VI-ReID), aimed at matching pedestrians matching across day and night.

\begin{figure}[!t]
\centering
\includegraphics[width=0.95\columnwidth]{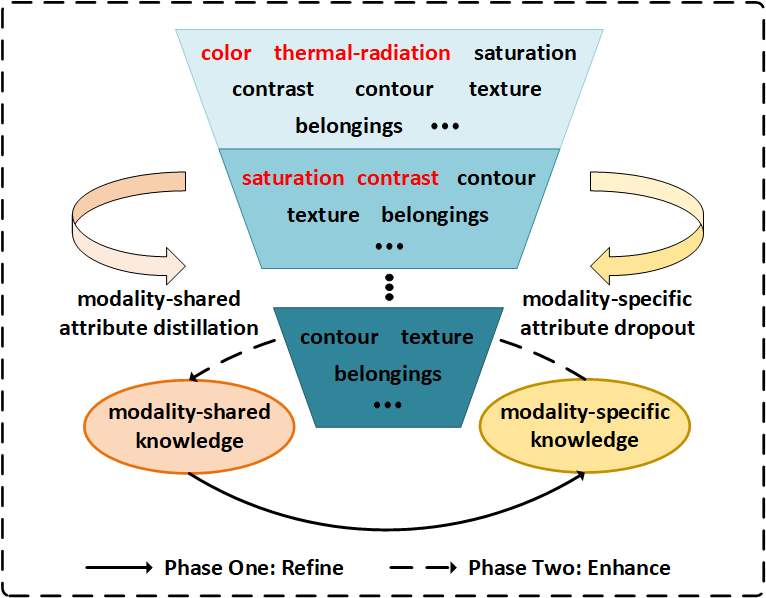}
\caption{Our ICRE is motivated by the idea that modality-specific attributes retained in shallow features can effectively enhance intra-class compactness after removing modality interference.}
\label{fig1}
\end{figure} 

VI-ReID aims to identity the same individual across different modalities \cite{r1, r14, r51} and its core challenge stems from the physical imaging differences between the two modalities: visible cameras capture detailed color and texture information through reflected short-wavelength light, while infrared sensors detect long-wavelength thermal radiation emitted by objects, resulting in grayscale representations of heat signatures. These modality-specific characteristics create substantial domain gaps in feature representation. 
In addition, challenges such as occlusions, pose changes and viewpoint variations further exacerbate the difficulty of this task, which collectively establish VI-ReID as one of the most demanding cross-modal matching problems in computer vision.

Recently, a variety of advanced methods have been introduced for VI-ReID, which can generally be divided into two main groups: the former group focuses on learning cross-modal representations directly from raw heterogeneous data, attempting to map them to a unified shared space \cite{r15, r9}, and the latter group aims to construct additional auxiliary samples to refine the feature space \cite{r29, r7}. Methods based on raw data rely on the powerful data analysis and induction capabilities of metric learning to seek identity-consistent representations in heterogeneous spectra for achieving feature-level alignment. However, due to the inherent resistance of heterogeneous images, there are still significant discrepancies between the extracted features. To this end, methods that utilize auxiliary data leverage techniques such as adversarial generative networks to generate cross-modal or intermediate samples, effectively narrowing the modality gap within the image domain.

While existing VI-ReID methods have made considerable progress by focusing on cross-modal shared features, they suffer from a critical limitation --
the neglect of modality-specific features that contain rich identity-related attributes (such as color, thermal radiation, and saturation).
These features are unique to each modality and contain substantial identity-related information, which, if properly harnessed, could significantly enhance recognition accuracy.
However, two major challenges emerge: (1) How to effectively extract modality-specific features? (2) How to refine these features to retain their latent identity-aware information while minimizing modality discrepancy?

Some methods extract modality-related features by adding independent network branches and utilize identity-aware knowledge through generating missing modality attributes \cite{r33, r17}. However, this manner not only increases the complexity of the model but also introduces additional noise disturbance in the generation process. We find that while shallow features of the modality-shared branch naturally preserve valuable modality-specific attributes (e.g., spectral signatures and texture patterns), these discriminative features are progressively attenuated in deeper layers through conventional fusion strategies, leaving only the common cues between the two modalities. Based on this, we propose a simple idea of distilling the identity-guiding knowledge embedded in shallow features and then enhance the discriminative capability of modality-invariant features, as illustrated in Figure~\ref{fig1}. It is worth noting that our approach has the following advantages: (1) no additional network branches are introduced; (2) potential identity interference that may arise during the generation of modality-specific attributes is avoided; (3) multi-scale semantics beneficial for identity recognition can be provided.

To this end, we propose an Identity Clue Refinement and Enhancement (ICRE) model for VI-ReID, which aggregates shallow features to capture modality-specific attributes and distills the embedded identity-guiding knowledge to further refine and enhance modality-invariant features. Since multiple shallow features have different semantic scales and modality attributes, they provide identity cues from different dimensions. Therefore, we first design a Multi-Perception Feature Refinement (MPFR) module to generate spatial masks that dynamically guide the fusion process of shallow features. MPFR adaptively reserves modality-specific attributes when constructing multi-granularity person descriptions. Then, to refine and enhance modality-invariant features, we introduce a Semantic Distillation Cascade Enhancement (SDCE) module, which consists of two cascaded attention blocks. The former refines shallow features to eliminate identity-irrelevant interference, while the latter distills and transfers identity-aware knowledge. In addition, to enhance the semantic modeling capability of the attention block, we add a Channel Modulation Factors (CMF) and a Joint Interaction Blocks (JIB) to the two residual paths. Finally, we design an Identity Clues Guided (ICG) Loss to facilitate cross-modal semantic transfer and effectively mitigate modality discrepancies.

Our main contributions are summarized below:
\begin{itemize}
\item[$\bullet$] We introduce an innovative Identity Clue Refinement and Enhancement (ICRE) network for VI-ReID, where discriminative identity-aware knowledge are systematically extracted from modality-specific shallow features and distills it into modality-invariant features.

\item[$\bullet$] We design a Multi-Perception Feature Refinement (MPFR) module to adaptively aggregate shallow features, enabling simultaneous capture of discriminative modality-specific attributes and complementary multi-scale semantics that are beneficial for identity matching.

\item[$\bullet$] We propose a Semantic Distillation Cascade Enhancement (SDCE) module that progressively extracts identity-aware knowledge from multi-scale shallow features and guide modality-invariant feature learning.

\item[$\bullet$] Experimental results show that our approach stays competitive when compared to other SOTA methods.
\end{itemize}

\section{Related Works}
\begin{figure*}[t]
\centering
\includegraphics[width=0.95\textwidth]{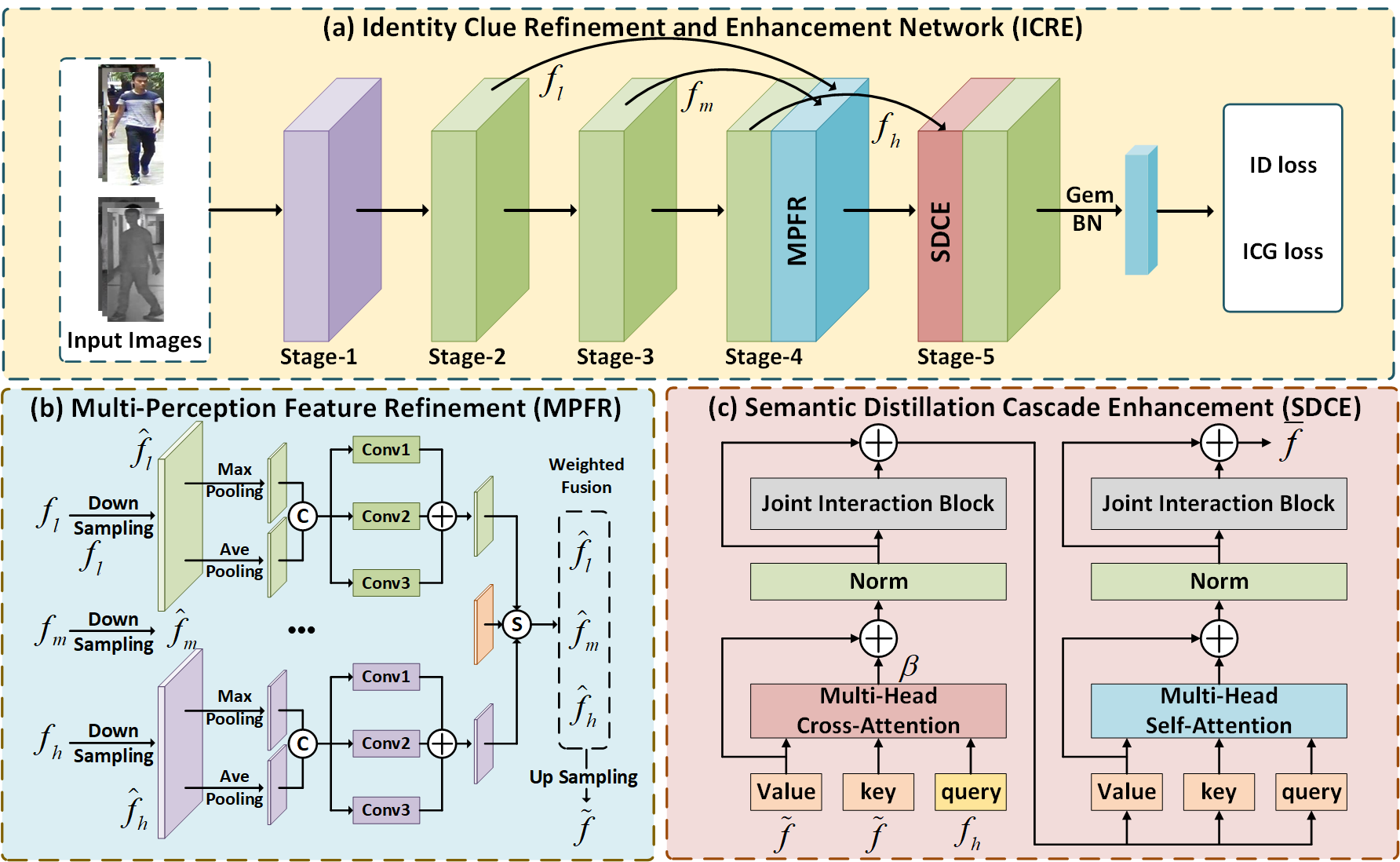}
\caption{(a) Overview of the proposed ICRE network, including a dual-stream backbone network, a Multi-Perception Feature Refinement (MPFR) module, and a Semantic Distillation Cascade Enhancement (SDCE) module. (b) and (c) illustrate the detailed processing flows of MPFR and SDCE, respectively. The network is ultimately constrained by both the identity (ID) loss and the identity clues guided (ICG) loss.}
\label{fig2}
\end{figure*}

\subsection{Visible Modality Person Re-identification}
Person ReID seeks to match the same individual from visible images taken by cameras with non-overlapping views \cite{r50, r49, r37, r38, r39, r46}. In recent years, researchers have paid more and more attention to designing effective metric losses to reduce intra-class variations while maximizing inter-class separability as much as possible \cite{r35}. However, feature extraction from full-body images often suffers from intra-modal discrepancies caused by factors such as partial occlusion, viewpoint changes, and clothing variations. Therefore, some methods adopt manual partitioning or attention mechanisms to capture fine-grained pedestrian descriptions, while integrating local and global information to enhance feature discriminability \cite{r34}. Moreover, to harness the cutting-edge semantic understanding of large language models (LLMs), Li et al. \cite{r26} applied the pre-trained CLIP model to person ReID, employing learnable ID-specific tokens to extract ambiguous text descriptions and guide visual feature learning. Person ReID methods have achieved remarkable success in a single visible modality, but it remains challenging to directly apply them to VI-ReID due to the substantial disparity between visible and infrared modalities.

\subsection{Visible-Infrared Person Re-identification}
VI-ReID is a task of cross-modal pedestrian retrieval that links daytime visible images with infrared images recorded at night \cite{r52, r53}. It is extremely difficult because of intra-modal variations (e.g., pose) and large inter-modal discrepancies caused by heterogeneous optical imaging processes. To address these challenges, some efforts have focused on aligning the representations of both modalities into a shared space for identity matching. Fang et al. \cite{r27} introduced learnable reference prototypes as latent depictions of body parts common to both modalities for achieving alignment of local features. In DMA\cite{r28}, a Dual Modality Transfer data augmentation technique was proposed, which applies color and brightness changes to image patches in the more color-sensitive HSV space for bridging the modality gap at the data level. DEEN \cite{r14} generated complementary and diversified feature embeddings, providing a more comprehensive description for identity recognition, and designed a multi-stage feature aggregation block to explore the potential spatial and channel correlations of features.

Additionally, there are some other works effectively refine the shared feature space by generating
auxiliary data to provide more specific identity clues. Qi et al. \cite{r29} designed an image generation network based on local cross-modal contrastive learning to generate cross-modal pseudo-images. Yu et al. \cite{r31} designed an auxiliary modality generator that extracts modality-common and identity-sensitive patterns from heterogeneous images to generate an intermediate modality for easy alignment. In SGIEL \cite{r30}, the model is guided to focus on modality-shared clues through semantic disentanglement and mutual information learning based on prior knowledge of human body shape.

\subsection{Extracting Modality-Specific Attributes for VI-ReID}
Modality-invariant representations play a crucial role in alleviating cross-modal discrepancies, while modality-related features significantly enhance intra-modal discriminative capabilities. Early approaches in this field deviated from the conventional paradigm of globally shared weights in feature extractors by incorporating modality-specific layers to facilitate more effective feature mining. These methodologies typically adopted a two-stage process: initially utilizing modality-specific layers to capture the unique discriminative semantics for each modality, followed by the application of modality-shared layers to extract cross-modality semantics, which substantially improves model performance \cite{r54, r55, r56, r57, r58}. Although these methods effectively highlighted the importance of modality-related features, their direct role in enhancing intra-class discrimination is not fully elucidated.

Recent researches have explored how to exploit modality-related features to enhance cross-modality recognition accuracy by introducing modality-specific network branches. Zhang et al.\cite{r32} developed an attention distillation module that facilities the transfer of useful style information from unimodal features to the cross-modal branch, mitigating intra-modal discriminative degradation. Lu et al.\cite{r33} refined and enhanced the complementarity of modality-invariant and modality-related features by computing their affinity relationship. Nevertheless, these methods typically require additional network branches for modality-specific attribute extraction, inevitably increasing both model parameter complexity and training process difficulty. To address the challenge of missing modality clues, alternative approaches employing generative strategies have been explored. Li et al.\cite{r48} proposed implementing modality-specific prototype memory banks for each modality, which generate modality-related attributes by calculating cosine similarity with modality-invariant features. Jiang et al.\cite{r62} proposed a novel framework incorporating two sets of learnable modality proxies and utilized Transformers to explore the relationships between original features and modality proxies, thereby compensating for missing modality-specific attributes. However, the modality prototypes or proxies used in these methods fail to capture modality-specific attributes unique to individual images and may introduce interference that leads to identity degradation.

In contrast to these approaches, our proposed method preserves modality-specific attributes through shallow feature aggregation. This strategy effectively mitigates the degradation of identity cues commonly seen in generative methods while maintaining model complexity at a manageable level.

\section{Methodology}
The comprehensive overview of our proposed Identity Clue Refinement and Enhancement (ICRE) model is provided in this section. Subsequently, we describe the structures of the proposed Multi-Perception Feature Refinement (MPFR) and Semantic Distillation Cascade Enhancement (SDCE) modules.

\subsection{Problem Formulation}
In the VI-ReID training set, visible and infrared images are represented as $V=\{x_{i}^{V},y_{i}^{V}\}_{i=\text{1}}^{{{N}_{V}}}$ and $I=\{x_{i}^{I},y_{i}^{I}\}_{i=\text{1}}^{{{N}_{I}}}$ respectively, where $x_{i}^{V}$ and $x_{i}^{I}$ represent the $i$-th visible and infrared samples, respectively. $y_{i}^{V}$ and $y_{i}^{I}$ are their corresponding identity labels, ${{N}_{V}}$ and ${{N}_{I}}$ represent the overall sample counts of the two modalities, respectively. Typically, a mini-batch comprises an equal number of visible and infrared images. We randomly sample $PK$ images from each modality, where $P$ and $K$ respectively denote the total number of classes and the count of images sampled per class. Moreover, the same $P$ identities are selected from both modalities, forming $PK$ cross-modal sample pairs fed into the feature extraction network. We follow the setup of \cite{r14, r18}, and use ResNet-50 as the backbone, whose first stage is set as the weight independent to capture discriminative semantics in each modality, while the weights of the subsequent four stages are shared, gradually filtering out modality-specific attributes to retain common cues. Furthermore, we replace the final average pooling operation of ResNet-50 with GeM pooling, which is between average and max pooling, and input the processed vector into BNneck to derive the final pedestrian embedding representation. During inference, the query and gallery sets are processed through their respective branches to extract feature representations. The affinity matrix between them is then computed and sorted according to the confidence score to obtain the retrieval results corresponding to the query image.


\subsection{Multi-Perception Feature Refinement}
Shallow features are refined by shared branches at different depths, thereby eliminating modality disparity to varying degrees. They can provide multi-granularity identity semantics and keep partial modality-related attributes for enhancing feature discriminability. Thus, we propose the Multi-Perception Feature Refinement module (MPFR), which adaptively selects the beneficial parts for cross-modal matching through a weighted fusion operation with spatial masks. We retain the output features of stages 2, 3 and 4, denoted as ${{f}_{l}}$, ${{f}_{m}}$ and ${{f}_{h}}$ respectively, and feed them into the our designed MPFR module for aggregation to obtain identity-guided feature $\tilde{f}$. The structure of MPFR is illustrated in Figure 2 (b), with its implementation details are described as follows.

\textbf{Feature scale alignment.} The output size of ResNet-50 varies at different stages, and needs to be consistent in spatial and channel dimensions. We employ ConvBlock, consisting of a convolutional layer, batch normalization, and the ReLU activation function, to adjust the features to a unified target scale, facilitating subsequent feature learning. The implementation process is as follows: 
\begin{equation}
\begin{aligned}
& {{{\hat{f}}}_{l}}=ConvBloc{{k}_{l}}({{f}_{l}}), \\ 
\end{aligned}
\end{equation}
where ${{\hat{f}}_{l}}$ is the determined size obtained through downsampling. The processing of ${{\hat{f}}_{m}}$ and ${{\hat{f}}_{h}}$ is similar to ${{\hat{f}}_{l}}$.

\textbf{Learning spatial mask.} Taking ${{\hat{f}}_{l}}$ as an example, we first apply max-pooling and average-pooling along the channel dimension to capture saliency and detail spatial information respectively. Then, concatenate the results to obtain a feature map ${{M}_{l}}\in {{\mathbb{R}}^{2\times H\times W}}$, where $H$ and $W$ represent the height and width of ${{\hat{f}}_{l}}$, respectively. Finally, we use three $3\times3$ dilated convolutional layers with different dilation rates (1, 2, 3) to realize the interaction between the two types of spatial information, resulting in the generation of a spatial mask. The specific generation process is as follows: 
\begin{equation}
\begin{aligned}
{{{M}'}_{l}}=(\varphi _{3\times 3}^{1}({{M}_{l}})+\varphi _{3\times 3}^{2}({{M}_{l}})+\varphi _{3\times 3}^{3}({{M}_{l}})),
\end{aligned}
\end{equation}
where $\varphi _{3\times 3}^{1}$ $\varphi _{3\times 3}^{2}$, and $\varphi _{3\times 3}^{3}$ are dilated convolutions with different dilation rates, ${{{M}'}_{l}}\in {{\mathbb{R}}^{1\times H\times W}}$ represents the spatial mask corresponding to ${{\hat{f}}_{l}}$, and the generation process of ${{{M}'}_{m}}$ and ${{{M}'}_{h}}$ similar to that of ${{{M}'}_{l}}$.

\textbf{Weighted feature refinement.} We utilize the Softmax function along the channel dimension to normalize the three spatial masks, obtaining spatial attention weights. 
\begin{equation}
\begin{aligned}
 & {{{{M}''}}_{i}}=\frac{exp({{{{M}'}}_{i}})}{\sum\limits_{j\in \{l,m,h\}}{exp({{{{M}'}}_{j}})}}, \\ 
\end{aligned}
\end{equation}
where ${{{M}''}_{i}}\in {{\mathbb{R}}^{1\times H\times W}}$ represents the normalized spatial attention weights. Then, the important shallow features are adaptively selected in the following manner:
\begin{equation}
\begin{aligned}
& \tilde{f}=ConvBlock(\sum\limits_{i\in \{l,m,h\}}{{{{{M}''}}_{i}}}\odot {{{\hat{f}}}_{i}}), \\ 
\end{aligned}
\end{equation}
where $\odot $ denotes element-wise multiplication implemented in broadcast mode. $ConvBlock(\cdot )$ is a block containing $1\times1$ convolution layer, which are used to restore the channel dimension of the fused features. $\tilde{f}$ is the final output of our proposed MPFR, and it has the same size as ${{f}_{h}}$. 

\subsection{Semantic Distillation Cascade Enhancement}
The identity-guided feature $\tilde{f}$ contains modality-invariant attributes that are beneficial for identifying pedestrian identities. However, compared to deep features, shallow features tend to introduce greater modality bias, which can interfere with identity-related cues. To distill identity discriminative knowledge from $\tilde{f}$ and guide the learning of modality-invariant features with relevant semantics, we design the Semantic Distillation Cascade Enhancement (SDCE) module. Benefiting from the powerful long-range modeling capability of the transformer, SDCE comprises two cascaded transformers. The former captures modality-related attributes most relevant to ${{f}_{h}}$ using cross-attention mechanism from $\tilde{f}$, and the latter transfers discriminative semantics inherent in the attributes to ${{f}_{h}}$ through a self-attention mechanism. We take the first transformer structure as an example to illustrate the specific implementation process.

\textbf{Transformer.} As shown in Figure~\ref{fig2} (c), we consider the spatial vectors corresponding to each pixel in the feature map as token embeddings. The generation process of Query ($Q$), Key ($K$) and Value ($v$) is as follows:
\begin{equation}
\small
\begin{aligned}
& Q=LN({{W}_{q}}({{f}_{h}})),K=LN({{W}_{k}}(\tilde{f})),V=LN({{W}_{v}}(\tilde{f})), \\ 
\end{aligned}
\end{equation}
where ${W}_{q}(\cdot)$, ${W}_{k}(\cdot)$ and ${W}_{v}(\cdot)$ are learnable weight matrices, and $LN(\cdot)$ represents the Layer Normalization operation.

The computation process of the Transformer Block output is as follows:
\begin{equation}
\small
\begin{aligned}
& out=MLP({V}')+{V}', \\ 
\end{aligned}
\end{equation}

\begin{equation}
\small
\begin{aligned}
& {V}'=LN(Attention(Q,K,V)+V), \\ 
\end{aligned}
\end{equation}
where ${V}'$ is the refined output after the attention layer, $Attention(\cdot)$ and $MLP(\cdot)$ refer to the multi-head attention mechanism and the multi-layer perceptron, respectively.

By modeling affinity semantics in spatial dimensions, we capture identity-guided knowledge most relevant to ${{f}_{h}}$ from $\tilde{f}$. However, unlike local features, which mainly exist in spatial vectors, identity-guided knowledge also occupies a significant proportion in the channel dimension. Therefore, we introduce a channel modulation factor and a joint interaction block to enhance the modeling capability of SDCE.

\textbf{Channel modulation factor.} We introduce a learnable vector $\beta \in \mathbb{R}^{1 \times C}$, which is initialized to 1, for the first residual block of the transformer. This vector is used to adaptively emphasize more important channels and suppress the interference of hidden noise. The residual calculation process is modified as follows:
\begin{equation}
\begin{aligned}
{V}'=LN(\beta \odot Attention(Q,K,V)+V),
\end{aligned}
\end{equation}
where $\odot $ denotes element-wise multiplication implemented in broadcast mode.

\textbf{Joint interaction block.} We modify the original MLP formed by a stack of a fully connected layer, a GELU activation function and another fully connected layer. Specifically, we add a branch that utilizes depthwise separable convolution to capture the spatial information of each channel.
\begin{equation}
\begin{aligned}
& weight=Reshape(DWConv(Reshape(F{{C}_{1}}(V')))), \\ 
\end{aligned}
\end{equation}
where $Reshape(\cdot )$ means the reshape operation, $DWConv(\cdot )$ is a depthwise separable convolution. This branch is employed to assess the importance of each pixel in this channel and enhance the original branch. The specific implementation is as follows:
\begin{equation}
\begin{aligned}
 & out=F{{C}_{3}}(GELU(F{{C}_{2}}({V}')weight)). \\ 
\end{aligned}
\end{equation}

The second transformer is similar to the first one, except that the inputs for computing the query, key and value are identical, and there is no channel modulation factors in the first residual block. The specific reasons will be explained in the experimental section.

\subsection{Joint training loss}
During the training phase, we impose different loss functions to constrain the sets ${{F}_{pool}}$ and ${{F}_{bn}}$ composed of feature vectors obtained after global pooling and BnNeck layer. For ${{F}_{pool}}$, we propose an Identity Clues Guide (ICG) loss to narrow the feature distance of different samples from the same identity across two modalities, while increasing the margin of features between different identity clusters. The features in $F_{bn}$ are fed into a classifier and optimized using the cross-entropy loss to encourage the network to learn modality-consistent identity semantics.

\textbf{Identity Clues Guide Loss.} For each mini-batch, we compute the visible, infrared and class-specific feature centers for each category as smoother supervisory signals to shrink or infer inter-modality samples. ICG can be expressed as:
\begin{equation}
\begin{aligned}
& {{L}_{ICG}}=\frac{1}{N}\sum\limits_{i=1}^{N}{{{[-{{\rho }_{1}}+||{{f}_{i}}-c_{{{y}_{i}}}^{{\bar{z}}}||]}_{+}}} \\ 
 & +\frac{2}{N(N+1)}\sum\limits_{\begin{smallmatrix} 
 i,j=1 \\ 
 \forall i\ne j 
\end{smallmatrix}}^{N}{{{[{{\rho }_{2}}-||{{f}_{i}}-{{c}_{{{y}_{j}}}}||]}_{+}},}
\end{aligned}
\end{equation}
where $N$ is the quantity of samples included in a mini-batch, ${{\rho }_{1}}$ and ${{\rho }_{2}}$ denote the margin parameters, ${{f}_{i}}$ expresses the $i$-th feature in ${{F}_{pool}}$, ${{[u]}_{+}}=max(u,0)$, and $||\cdot ||$ means L2 Norm operation. ${{y}_{i}}$ and $\bar{z}\in \{V,I\}$ represent the identity and another modal label of ${{f}_{i}}$ respectively. $c_{yi}^{{\bar{z}}}$ and ${{c}_{{{y}_{j}}}}$ denote the corresponding intra-modality and intra-identity feature centers respectively.

The total loss used for training can be written as follows:
\begin{equation}
\begin{aligned}
L={{L}_{ce}}+\lambda {{L}_{ICG}},
 \end{aligned}
\end{equation}
where ${{L}_{ce}}$ is the cross-entropy loss and $\lambda $ denotes the loss balance hyperparameter.

\begin{table*}[]
\begin{center}
\caption{Performance comparison with other SOTA approaches on SYSU-MM01. The top-performing method is marked in bold, while the second-best result is indicated with underlined fonts. CMC (\%) and mAP (\%) are reported}
\label{tab1}
\resizebox{1.8\columnwidth}{!}{
\begin{tabular}{cc|cccc|cccc}
\hline
\multicolumn{2}{c|}{Settings} &
\multicolumn{4}{c|}{All-Search} &
\multicolumn{4}{c}{Indoor-Search} \\
\hline
\multicolumn{1}{c|}{\multirow{2}{*}{Method}} &
\multirow{2}{*}{Venue} &
\multicolumn{2}{c|}{Single-Shot} &
\multicolumn{2}{c|}{Multi-Shot} &
\multicolumn{2}{c|}{Single-Shot} &
\multicolumn{2}{c}{Multi-Shot} \\ \cline{3-10} 
\multicolumn{1}{c|}{} & & R1  & \multicolumn{1}{c|}{mAP} & R1 & mAP & R1 & \multicolumn{1}{c|}{mAP} & R1 & mAP \\ \hline
\multicolumn{1}{l|}{MAUM \cite{r9}} & \multicolumn{1}{l|}{CVPR 22} & 61.59 & \multicolumn{1}{c|}{59.96} &- & - &  67.07 & \multicolumn{1}{c|}{73.58} & - & - \\
\multicolumn{1}{l|}{DART \cite{r22}} & \multicolumn{1}{l|}{CVPR 22} & 68.72 & \multicolumn{1}{c|}{66.29} &- & - &  72.52 & \multicolumn{1}{c|}{78.17} & - & - \\
\multicolumn{1}{l|}{DCLNet \cite{r20}} & \multicolumn{1}{l|}{MM 22} & 70.79 & \multicolumn{1}{c|}{65.18} &- & - &  73.51 & \multicolumn{1}{c|}{76.80} & - & - \\
\multicolumn{1}{l|}{MSCLNet  \cite{r11}} & \multicolumn{1}{l|}{ECCV 22} & 76.99 & \multicolumn{1}{c|}{71.64} &- & - & 78.49 & \multicolumn{1}{c|}{81.17} &- & - \\
\multicolumn{1}{l|}{KMDL \cite{r10}} & \multicolumn{1}{l|}{MM 22} & 77.61 & \multicolumn{1}{c|}{74.24} &- & - & 80.76 & \multicolumn{1}{c|}{69.31} &-& - \\
\multicolumn{1}{l|}{CMTR  \cite{r13}} & \multicolumn{1}{l|}{TMM 23} & 65.45 & \multicolumn{1}{c|}{62.90} &71.99 & 57.07 & 71.46  & \multicolumn{1}{c|}{76.67} &80.00 & 69.49 \\
\multicolumn{1}{l|}{CAL  \cite{r65}} & \multicolumn{1}{l|}{ICCV 23} & 74.66 & \multicolumn{1}{c|}{71.73} & 77.05 & 64.86 & 79.69 & \multicolumn{1}{c|}{83.68} & 86.97 & 78.51 \\
\multicolumn{1}{l|}{PMT  \cite{r23}} & \multicolumn{1}{l|}{AAAI 23} & 67.53 & \multicolumn{1}{c|}{66.42} &- & - & 71.66  & \multicolumn{1}{c|}{ 76.52} &- & - \\
\multicolumn{1}{l|}{MRCN  \cite{r12}} & \multicolumn{1}{l|}{AAAI 23} & 68.90 & \multicolumn{1}{c|}{65.50} &- & - & 76.00  & \multicolumn{1}{c|}{ 79.80} &- & - \\
\multicolumn{1}{l|}{DEEN  \cite{r14}} & \multicolumn{1}{l|}{CVPR 23} &74.70 & \multicolumn{1}{c|}{71.80} &- & - & 80.30  & \multicolumn{1}{c|}{83.30} &- & - \\
\multicolumn{1}{l|}{SAAL  \cite{r27}} & \multicolumn{1}{l|}{ICCV 23} & 75.90 & \multicolumn{1}{c|}{\textbf{77.03}} & 82.86 & \textbf{82.39} & 83.20 & \multicolumn{1}{c|}{88.01} & 90.73 & \textbf{91.30} \\
\multicolumn{1}{l|}{MUN  \cite{r31}} & \multicolumn{1}{l|}{ICCV 23} & 76.24 & \multicolumn{1}{c|}{73.81} & - & - & 79.42 & \multicolumn{1}{c|}{82.06} & - & - \\
\multicolumn{1}{l|}{SEFL  \cite{r30}} & \multicolumn{1}{l|}{CVPR 23} & 77.12 & \multicolumn{1}{c|}{72.33} & - & - & 82.07 & \multicolumn{1}{c|}{82.95} & - & - \\
\multicolumn{1}{l|}{PartMix  \cite{r15}} & \multicolumn{1}{l|}{CVPR 23} &77.78 & \multicolumn{1}{c|}{74.62} &80.54 & 69.84 & 81.52  & \multicolumn{1}{c|}{84.38} &87.99 & 79.95 \\
\multicolumn{1}{l|}{PMWGCN\cite{r60}} & \multicolumn{1}{l|}{TIFS 24} & 66.82 & \multicolumn{1}{c|}{64.88} &- & - & 72.64 & \multicolumn{1}{c|}{76.19} &- & - \\
\multicolumn{1}{l|}{DARD  \cite{r16}} & \multicolumn{1}{l|}{TCSVT 24} & 69.33 & \multicolumn{1}{c|}{65.65} &79.63 & 60.30 & 77.21 & \multicolumn{1}{c|}{81.91} &86.63 & 75.28 \\
\multicolumn{1}{l|}{MFCS  \cite{r17}} & \multicolumn{1}{l|}{TMM 24} & 70.59 & \multicolumn{1}{c|}{67.49} & - & - &  75.98 & \multicolumn{1}{c|}{ 81.55} & - & - \\
\multicolumn{1}{l|}{CAJ  \cite{r18}} & \multicolumn{1}{l|}{TPAMI 24} & 71.48 & \multicolumn{1}{c|}{68.15} &- & - &  78.36 & \multicolumn{1}{c|}{81.98} &- & - \\
\multicolumn{1}{l|}{CSC-Net\cite{r59}} & \multicolumn{1}{l|}{TCSVT 24} & 72.70 & \multicolumn{1}{c|}{69.58} &78.14 &64.99 & 78.56 & \multicolumn{1}{c|}{82.12} &85.45 &77.67 \\
\multicolumn{1}{l|}{DMPF \cite{r19}} & \multicolumn{1}{l|}{TNNLS 24} &76.41 & \multicolumn{1}{c|}{71.55} &- & - & 82.29  & \multicolumn{1}{c|}{84.94} &- &- \\ 
\multicolumn{1}{l|}{HOS-Net \cite{r36}} & \multicolumn{1}{l|}{AAAI 24} &75.60 & \multicolumn{1}{c|}{74.20} &- & - &84.20  & \multicolumn{1}{c|}{86.70} &- & - \\ 
\multicolumn{1}{l|}{DNS  \cite{r67}} & \multicolumn{1}{l|}{ECCV 24} & 77.27 & \multicolumn{1}{c|}{74.35} & - & - & 84.21 & \multicolumn{1}{c|}{86.83} & - & - \\
\multicolumn{1}{l|}{WRIM-Net  \cite{r66}} & \multicolumn{1}{l|}{ECCV 24} & 77.40 & \multicolumn{1}{c|}{75.40} & \underline{83.20} & 71.10 & \textbf{86.20} & \multicolumn{1}{c|}{\textbf{88.10}} & \textbf{92.10} & \underline{84.60} \\
\multicolumn{1}{l|}{DSAF  \cite{r68}} & \multicolumn{1}{l|}{TMM 25} & 76.65 & \multicolumn{1}{c|}{73.24} & - & - & 83.48 & \multicolumn{1}{c|}{83.78} & - & - \\
\multicolumn{1}{l|}{MSCMNet  \cite{r69}} & \multicolumn{1}{l|}{PR 25} & \underline{78.53} & \multicolumn{1}{c|}{74.20} & - & - & 83.00 & \multicolumn{1}{c|}{85.54} & - & - \\
\hline
\multicolumn{1}{l|}{AlignGAN \cite{r3}} & \multicolumn{1}{l|}{ICCV 19} & 42.40 & \multicolumn{1}{c|}{40.70} & 51.50 & 33.90 & 45.90 & \multicolumn{1}{c|}{54.30} & 57.10 & 45.30 \\
\multicolumn{1}{l|}{cm-SSFT \cite{r33}} & \multicolumn{1}{l|}{CVPR 20} & 61.60 & \multicolumn{1}{c|}{63.20} & 63.40 & 62.00 & 70.50 & \multicolumn{1}{c|}{72.60} & 73.00 & 72.41 \\
\multicolumn{1}{l|}{DML \cite{r32}} & \multicolumn{1}{l|}{TCSVT 22} & 58.40 & \multicolumn{1}{c|}{56.10} & 62.20 & 49.60 & 62.40 & \multicolumn{1}{c|}{69.50} & 66.40 & 60.00 \\
\multicolumn{1}{l|}{ACD\cite{r61}} & \multicolumn{1}{l|}{TIFS 24} & 74.44 & \multicolumn{1}{c|}{71.17} &80.45 &66.90 & 78.98 & \multicolumn{1}{c|}{82.75} &86.70 &78.69 \\
\multicolumn{1}{l|}{\textbf{ICRE (ours)}} & \multicolumn{1}{l|}{-} &\textbf{80.41} & \multicolumn{1}{c|}{\underline{76.96}} &\textbf{86.54} & \underline{72.68} & \underline{85.04}  & \multicolumn{1}{c|}{87.09} &\underline{91.62} &82.94 \\
\hline
\end{tabular}}
\end{center}
\end{table*}

\section{Experiments}

\begin{table}
\centering
\caption{Comparison with other SOTA approaches on RegDB. The top-performing method is marked in bold, while the second-best result is indicated with underlined fonts}
\label{tab2}
\resizebox{0.95\columnwidth}{!}{
\begin{tabular}{l|cc|cc}
\hline
\multicolumn{1}{c|}{\multirow{2}{*}{Method}} & \multicolumn{2}{c|}{VIS to IR} & \multicolumn{2}{c}{IR to VIS} \\ \cline{2-5} 
                        & R1             & mAP              & R1            & mAP              \\ \hline
MAUM\cite{r9}                   & 83.39              & 78.75           & 81.07             & 78.89            \\
DART\cite{r22}                   & 83.60              & 75.67           & 81.97             & 73.78            \\
DCLNet\cite{r20}         & 81.20             & 74.30          & 78.00             & 70.60            \\
MSCLNet\cite{r11}                  & 84.17              &  80.99            & 83.86             &  78.31           \\
KMDL\cite{r10}    & 85.08              & 81.01           & 84.42             & 81.28            \\
CMTR\cite{r13}                  & 88.11              & 81.66            & 84.92             & 80.79            \\
CAL\cite{r65} & 94.51 & 88.67 & \underline{93.64} & 87.61 \\
PMT\cite{r23}          & 84.83             & 76.55           & 84.16             & 75.13            \\
MRCN\cite{r12}          & 91.40              & 84.60           & 88.30             & 81.90            \\
DEEN\cite{r14}                    & 91.10              & 85.10            & 89.50             & 83.40            \\
SAAI\cite{r27} & 91.07 & \textbf{91.45} & 92.09 & \textbf{92.01} \\
MUN\cite{r31} & \textbf{95.19} & 87.15 & 91.86 & 85.01 \\
SEFL\cite{r30} & 92.18 & 86.59 & 91.07 & 85.23 \\
PartMix\cite{r15}   & 85.66              & 82.27            & 84.93             & 82.52            \\
PMWGCN\cite{r60}            & 90.61             & 84.53           & 88.77             & 81.61           \\  
DARD\cite{r16}                   & 86.19            & 85.39            & 85.53             & 85.09           \\
MFCS\cite{r17}       & 85.34              & 76.39            & 83.88             & 75.16            \\ 
CAJ\cite{r18}                  & 85.69               & 79.70            & 84.88             & 78.55            \\ 
CSC-Net\cite{r59}                 & 90.97             & 86.43            & 89.42             &85.65           \\  
DMPF\cite{r19}                  & 88.83              & 81.02            & 88.88             & 81.86           \\  
HOS-Net \cite{r36}          & \underline{94.70}            & 90.40           & 93.30             & 89.20 \\ 
DNS\cite{r67} & 93.01 & 88.56 & 93.48 & 88.10 \\
WRIM-Net\cite{r66} & 94.50 & \underline{90.50} & \textbf{93.70} & \underline{89.70} \\
DSAF\cite{r68} & 93.25 & 87.17 & 92.62 & 86.37 \\
MSCMNet\cite{r69} & 90.40 & 81.20 & 87.70 & 78.20 \\
\hline
AlignGAN\cite{r3}                & 57.90              & 53.60            & 56.30             & 53.40            \\
cm-SSFT \cite{r33}   & 72.30             & 72.90           & 71.00             & 71.70           \\  
DML \cite{r32}   & 77.60             & 74.30           & 77.00             & 83.60           \\  
ACD\cite{r61}   & 85.10             & 81.32           & 85.24             & 82.36            \\  
\textbf{ICRE (ours)}            & 91.56             & 85.63          & 90.16            & 84.32 \\
\hline
\end{tabular}}
\end{table}

\subsection{Datasets and Evaluation Protocols}
\textbf{Datasets.}
We evaluate the effectiveness of our model on three VI-ReID benchmark datasets: SYSU-MM01 \cite{r1}, LLCM \cite{r14} and RegDB \cite{r25}. SYSU-MM01 contains 30,071 visible images and 15,792 infrared images from 491 unique identities. The dataset is divided into training and testing sets based on specific IDs, and provides both all-search and indoor-search evaluation modes. The LLCM dataset considers challenges such as low lighting conditions and clothing variations, comprising 16,946 visible images and 13,975 infrared images from 713 different identities. The dataset is split into training and testing sets in a 2:1 ratio. The RegDB dataset is captured by a surveillance system with dual cameras, comprising 412 identities. Half of these identities are used for training, while the other half are used for testing. Each identity has 10 pairs of cross-modal images.

\textbf{Evaluation Protocols}
We employ standard Cumulative Matching Characteristics (CMC) and mean Average Precision (mAP) as our evaluation metrics.

\subsection{Implementation details}
We implement our proposed ICRE using PyTorch and train it on a single RTX 4090 GPU. In the training phase, the size of all input images are fixed to $384\times 192$, and data augmentation techniques such as random erasing, grayscale, random horizontal flip and channel augmentation are selected to increase the richness of the data. Notably, channel augmentation is exclusively applied to visible images with an application probability 50\%. We adopt AGW \cite{r21} as our model baseline, selecting 48 images randomly for each iteration to form a mini-batch, with half from the visible modality and the other half from the infrared modality. To maintain balanced training, these mini-batches are evenly sampled from 6 randomly selected identities. We employ a gradual warm-up approach to increment the learning rate from 0.01 to 0.1 during the first 10 epochs. Subsequently, at the 30th, 90th and 120th epochs, we adjust the learning rates to 0.01, 0.001 and 0.0001. We choose SGD as our optimizer, and the momentum parameter is set to 0.9. The total training duration consists of 150 epochs. Through ablation experiments, we verify that setting the hyperparameter $\lambda$ in the total loss function to 1 and applying the channel modulation factor only in the first Transformer block yields the best performance.

\subsection{Comparison with State-of-the-Art Methods}
\textbf{Comparisons on SYSU-MM01 and RegDB Datasets.}
We evaluate the performance of our proposed ICRE against several SOTA VI-ReID methods on the SYSU-MM01 and RegDB datasets, including MAUM\cite{r9}, DART\cite{r22}, DCLNet\cite{r20}, MSCLNet\cite{r11}, KMDL\cite{r10}, CMTR\cite{r13}, CAL\cite{r65}, PMT\cite{r23}, MRCN\cite{r12}, DEEN\cite{r14}, SAAL\cite{r27}, MUN\cite{r31}, SEFL\cite{r30}, PartMix\cite{r15}, PMWGCN\cite{r60}, DARD\cite{r16}, MFCS\cite{r17}, CAJ\cite{r18}, CSC-Net\cite{r59}, DMPF\cite{r19}, HOS-Net\cite{r36}, DNS\cite{r67}, WRIM-Net\cite{r66}, DSAF\cite{r68}, MSCMNet\cite{r69}, AlignGAN\cite{r3}, cm-SSFT\cite{r33}, DML\cite{r32} and ACD\cite{r61}. Among them, all methods follow the paradigm of not employing re-ranking.

From Table \uppercase\expandafter{\romannumeral1}, we find that ICRE achieves the best Rank-1 accuracy under the most challenging All-Search setting, reaching 80.41\% and 86.54\% in the Single-Shot and Multi-Shot evaluation modes, respectively. It also achieves second-best or competitive performance on other evaluation metrics. Even compared to KMDL \cite{r10} and DMPF \cite{r19}, which incorporate keypoint and skeleton priors respectively, our method shows a significant advantage while maintaining model simplicity. Our ICRE yields slightly lower mAP scores compared with SAAI\cite{r27}, mainly because employs an affinity inference matrix during testing to better balance the distance metrics between hard and easy samples within the same modality. However, this approach does not conform to the standard evaluation protocol and introduces unfair factors. Moreover, our ICRE performs slightly worse than WRIM-Net\cite{r66} under the Indoor-Search setting, mainly because WRIM-Net extracts wide-ranging information across spatial and channel dimensions, enabling it to capture local cues more effectively in the relatively simple indoor environment. The above results indicate that our ICRE effectively enhances the accuracy of person recognition by leveraging modality-specific attributes and multi-scale clues from shallow features.


It can be observed from Table \uppercase\expandafter{\romannumeral2} that our ICRE achieves a Rank-1 retrieval accuracy of 90.56\% in VIS to IR search mode and 91.16\% in IR to VIS search mode, demonstrating competitive performance compared with some methods. However, our method is slightly inferior to methods such as HOS-Net \cite{r36} and WRIM-Net \cite{r66}. The main reason is that the infrared images in RegDB belong to the far-infrared spectrum, which exhibits a greater modality gap from visible images compared to the near-infrared images in datasets like SYSU-MM01 or LLCM. As a result, the model needs to rely more heavily on modality-invariant structural cues such as body proportions. HOS-Net \cite{r36} constructs high-order structures using short-range and long-range features, while WRIM-Net \cite{r66} mines large-scale multi-dimensional information, enabling both to capture rich local cues and achieve superior performance. Moreover, due to the larger modality style discrepancy in the RegDB dataset, our proposed ICRE faces greater challenges in filtering and utilizing useful information from modality-specific attributes, resulting in relatively limited performance improvement. In future work, we aim to explore solutions to this issue by focusing on style normalization techniques.

As demonstrated in Tables \uppercase\expandafter{\romannumeral1} and \uppercase\expandafter{\romannumeral2}, we further compare our method with several related approaches, including AlignGAN [52], ACD [53], cm-SSFT [10] and DML [32]. Methods [10] and [32] significantly increase the model complexity by introducing additional modality-specific branches to extract modality-specific attributes. AlignGAN and ACD generate missing modality attributes at the image or feature level, but the unpredictability of the generation process may introduce identity interference. In contrast, our proposed ICRE effectively preserves modality-specific attributes by aggregating shallow features, achieving SOTA results while avoiding the introduction of excessive parameters or generative noise.


\textbf{Comparisons on LLCM Dataset.} As shown in Table \uppercase\expandafter{\romannumeral3}, we compare our method with DDAG \cite{r4}, AGW \cite{r21}, LbA \cite{r5}, DCLNet \cite{r20}, CAJ\cite{r18}, PMT\cite{r23}, DART\cite{r22}, DEEN\cite{r14}, DNS\cite{r67}, WRIM-Net\cite{r66}, DSAF\cite{r68}, MSCMNet\cite{r69} and HOS-Net\cite{r36} on the recently proposed LLCM dataset, which encompasses complex conditions such as varying lighting. We observe that our method achieves the highest Rank-1 accuracy of 58.4\% under the IR to VIS search mode and also attains the second-best performance across other evaluation metrics. It achieves comparable performance to WRIM-Net, which captures wide-ranging information through multi-dimensional interactions to enhance robustness against noise such as illumination and pose variations. This indicate that our ICRE effectively leverages modality-specific knowledge to capture more pedestrian-relevant cues, thereby better handling sample diversity.

\textbf{Generalization Capability Evaluation on MSVR310 Dataset.} To comprehensively evaluate the generalization ability of our proposed ICRE in cross-modal tasks, we conduct relevant experiments on the vehicle cross-modal dataset MSVR310\cite{r64} and compare it with several open-source methods, including DDAG \cite{r4}, LbA \cite{r5}, DSCNet \cite{r63}, and DEEN\cite{r14}. As shown in Table \uppercase\expandafter{\romannumeral4}, our method consistently outperforms existing methods across all evaluation metrics. These results indicate that the ICRE framework effectively leverages modality-specific attributes encoded in shallow features, thereby significantly enhancing intra-class discriminability in cross-modal matching tasks.

\begin{table}
\centering
\caption{Comparison with other SOTA approaches on LLCM. The top-performing method is marked in bold, while the second-best result is indicated with underlined fonts}
\label{tab3}
\resizebox{.85\columnwidth}{!}{
\begin{tabular}{l|cc|cc}
\hline
\multicolumn{1}{c|}{\multirow{3}{*}{Method}} & \multicolumn{4}{c}{LLCM} \\ \cline{2-5} 
 & \multicolumn{2}{c|}{VIS to IR} & \multicolumn{2}{c}{IR to VIS} \\ \cline{2-5} 
                        & R1             & mAP              & R1            & mAP              \\ \hline
DDAG \cite{r4}            & 48.0              & 52.3           & 40.3             & 48.4           \\
AGW\cite{r21}                   & 51.5             & 55.3            & 43.6              & 51.8            \\
LbA\cite{r5}        & 50.8             & 55.6           &  43.8               & 53.1         \\
DCLNet\cite{r20}                   & 52.2             & 55.7            & 44.5             & 50.9          \\
CAJ\cite{r18}                   & 56.5             & 59.8            & 48.8              & 56.6           \\
PMT\cite{r23}                   & 58.4             & 62.1           & 49.9             & 57.2           \\
DART\cite{r22}                   & 60.4             & 63.2            & 52.2              & 59.8           \\
DEEN\cite{r14}                   & 62.5            & 65.8            & 54.9             & 62.9          \\ 
HOS-Net \cite{r36}                 & 64.9             & 67.9            & 56.4            & 63.2           \\
DNS\cite{r67}                   & 66.0            & 68.6            & \underline{57.5}             & 64.1         \\ 
WRIM-Net\cite{r66}                   & \textbf{67.7}            & \textbf{69.2}            & 58.4             & \textbf{64.8}         \\
DSAF\cite{r68}                   & 65.4            & 68.2            & 57.3             & 64.3         \\
MSCMNet\cite{r69}                   & 64.3            & 66.6            & 56.5             & 63.5         \\  
\hline
\textbf{ICRE (ours)}            & \underline{67.2}              & \underline{69.1}           & \textbf{58.4}             & \underline{64.7} \\
\hline
\end{tabular}}
\end{table}

\begin{table}
\centering
\caption{The Generalization Verification of ICRE on Vehicle Cross-Modal Dataset MSVR310}
\label{tab4}
\resizebox{.85\columnwidth}{!}{
\begin{tabular}{l|cc|cc}
\hline
\multicolumn{1}{c|}{\multirow{3}{*}{Method}} & \multicolumn{4}{c}{MSVR310} \\ \cline{2-5} 
 & \multicolumn{2}{c|}{VIS to IR} & \multicolumn{2}{c}{IR to VIS} \\ \cline{2-5} 
                        & R1             & mAP              & R1            & mAP              \\ \hline
DDAG \cite{r4}            & 37.90              & 23.14          & 39.13             & 22.49           \\
LbA \cite{r5}             & 32.32             & 18.67          & 31.18              & 18.13        \\
DSCNet \cite{r63}         & 48.81             & 29.29          & 49.36              & 29.79        \\
DEEN \cite{r14}           & 46.73             & 28.68          & 45.89              & 28.54        \\ 
\textbf{ICRE (ours)}      & \textbf{51.53}    & \textbf{31.89} & \textbf{50.13}     & \textbf{31.23} \\
\hline
\end{tabular}}
\end{table}

\begin{table}[!t]    
\begin{center}    
\caption{Evaluate the impact of each individual component of the proposed method on SYSU-MM01}   
\label{tab5}    
\renewcommand{\arraystretch}{1.5} 
\resizebox{0.99\columnwidth}{!}{
\begin{tabular}{c|c|c|c|c|c|c}    
\hline    
\multicolumn{5}{c|}{\textbf{Settings}} & \multicolumn{2}{c}{\textbf{SYSU-MM01}} \\ \hline    
\multicolumn{1}{c|}{BASE} & \multicolumn{1}{c|}{MPDR} & \multicolumn{1}{c|}{SDCE} & \multicolumn{1}{c|}{$L_{TRI}$} & \multicolumn{1}{c|}{$L_{ICG}$} & \multicolumn{1}{c|}{R1} & \multicolumn{1}{c}{mAP} \\ \hline    
$\checkmark$ &$\times$ &$\times$ &$\checkmark$ &$\times$ &70.21 &68.48 \\ \hline
$\checkmark$ &$\times$ &$\times$ &$\times$ &$\checkmark$ &72.03 &69.54 \\ \hline 
$\checkmark$ &$\checkmark$ &$\times$ &$\checkmark$ &$\times$ &76.22 &72.66 \\ \hline
$\checkmark$ &$\checkmark$ &$\times$ &$\times$ &$\checkmark$ &77.51 &74.16 \\ \hline    
$\checkmark$ &$\checkmark$ &$\checkmark$ &$\checkmark$ &$\times$ &78.90 &74.94 \\ \hline 
$\checkmark$ &$\checkmark$ &$\checkmark$ &$\times$ &$\checkmark$ &\textbf{80.41} &\textbf{76.96} \\ \hline    
\end{tabular}}    
\end{center}    
\end{table}

\subsection{Ablation studies}
\textbf{Effectiveness of each components.} To validate the effectiveness of each component in our proposed ICRE, we perform ablation experiments on the SYSU-MM01 dataset and the results are presented in Table V. To effectively aggregate shallow features with inconsistent scales, we design MPFR to adjust the feature scales and selectively enhance discriminative regions, thereby facilitating the integration of multi-scale identity clues and preserving modality-specific attributes. We observe that when MPDR is added to the baseline model, the Rank-1 accuracy and mAP score increased by 6.01\% and 4.18\%, respectively. This indicates that MPDR can enhance the discriminative power of pedestrian representations by generating spatial masks to remove redundant information and noise interference in modality-specific attributes. Furthermore, this highlights the crucial role of modality-specific attributes in improving intra-class discriminability.

In addition, we design SDCE to extract identity-discriminative knowledge from modality-specific attributes through semantic filtering and contextual association modeling, thereby guiding the learning of modality-invariant features. After introducing SDCE, the Rank-1 accuracy and mAP score are further improved to 78.90\% and 74.94\%, respectively. This demonstrates that SDCE enhances the semantic modeling capability of the Transformer by incorporating Channel Modulation Factors and Joint Interaction Blocks. It effectively preserves the core identity semantics within each modality while mitigating modality gap interference, thereby improving the representation consistency between visible and infrared features.

\begin{table}[]
\begin{center}
\caption{The influence of utilizing shallow features from different stages in the MPDR. Where "stage(1,2,3,4)" refers to retaining the output from Stage-1 to Stage-4. The superscript "*" indicates the use of the ICG loss, while the absence of a symbol indicates the use of the Triplet loss.}
\label{tab6}
\renewcommand{\arraystretch}{1.2} 
\resizebox{1\columnwidth}{!}{
\begin{tabular}{l|cccc}
\hline
Setting               &  Rank-1 & Rank-10 & Rank-20 & mAP   \\ \hline
Stage (1,2,3,4) & 75.22 & 96.92 & 99.17 & 72.18 \\
Stage (2,3,4) & 78.90 & 97.73 & 99.29 & 74.94 \\
Stage (3,4) & 77.42 & 97.09 & 99.21 & 74.03 \\
Stage (4) & 76.18 & 97.46 & 99.35 & 72.55 \\ \hline
Stage (1,2,3,4)\textsuperscript{*}         & 79.85  & 97.93   & 99.32   & 76.21 \\
Stage (2,3,4)\textsuperscript{*}          & \textbf{80.41}  & \textbf{98.19}   & \textbf{99.55}   & \textbf{76.96} \\ 
Stage (3,4)\textsuperscript{*}          & 80.12 & 98.06   & 99.44   & 76.53 \\
Stage (4)\textsuperscript{*}         & 79.74  & 97.91   & 99.16   & 76.32 \\ \hline
\end{tabular}}
\end{center}
\end{table}

\begin{table}[]    
\begin{center}    
\caption{Validate the effectiveness of SDCE in refining and enhancing features under different settings}   
\label{tab7}    
\renewcommand{\arraystretch}{1.3} 
\resizebox{1\columnwidth}{!}{
\begin{tabular}{ccccc|cc}    
\hline      
$Atte{{n}_{cro}}$ &$Atte{{n}_{self}}$ &$JIB$ &$\beta_1$ &$\beta_2$ &R-1 &mAP \\ \hline    
$\checkmark$ & & & & &78.37 &74.36 \\
$\checkmark$ &$\checkmark$ & & & &78.62 &74.87 \\
$\checkmark$ &$\checkmark$ &$\checkmark$ & & &79.93 &75.68 \\
$\checkmark$ &$\checkmark$ &$\checkmark$ &$\checkmark$ & &\textbf{80.41} &\textbf{76.96} \\
$\checkmark$ &$\checkmark$ &$\checkmark$ &$\checkmark$ &$\checkmark$ &79.89 &75.32 \\\hline   
\end{tabular}}    
\end{center}    
\end{table}

\begin{table*}[t]
\begin{center}
\caption{The performance of our method, in terms of parameter count, FLOPs, inference times, training times and Rank-1 accuracy, is compared with other methods on the SYSU-MM01}
\label{tab8}
\renewcommand{\arraystretch}{1.2} 
\resizebox{1.5\columnwidth}{!}{
\begin{tabular}{c|c|c|c|c|c}
\hline
Method            &Params (M) &FLOPs (G)  &Inference time (S) &Training time (H)  &R-1 \\ \hline
CAJ \cite{r18}          & 30.8  & 4.3  & 44.5 & 3.5 & 71.48 \\
DEEN \cite{r14}          & 41.2  & 21.8  & 63.8 & 10.5 & 74.70 \\
HOS-Net \cite{r36}        & 83.4  & 14.3  & 50.7 & 32.4 & 75.60 \\
ICRE (Ours)             & 38.0   & 6.4    & 49.1 & 7.7  & \textbf{80.41}\\ \hline
\end{tabular}}
\end{center}
\end{table*}

Meanwhile, we find that replacing $L_{TRI}$ with {$L_{ICG}$ can improve the performance of the proposed components under different combinations. This indicates that feature centers provide a more representative and generalizable identity representation in both modalities compared to using the hardest or easiest samples as distance constraints. The intra-modality and intra-identity feature centers in {$L_{ICG}$ are respectively used to convey cross-modal knowledge and enhance the discriminability between identities, effectively alleviating the clue degradation caused by modality discrepancy. Additionally, the hyperparameter ${\rho }_{1}$ in $L_{TRI}$ facilitates the learning of a diverse sample space, thereby enhancing adaptability to hard-to-distinguish samples.

\begin{table}[]    
\begin{center}    
\caption{Influence of batch size on model performance}   
\label{tab9}    
\renewcommand{\arraystretch}{1.2} 
\resizebox{1\columnwidth}{!}{
\begin{tabular}{ccc|cc} 
\hline      
Batch Size & ID count & Images per ID & Rank-1 & mAP \\ \hline    
64 & 8 & 8 & 79.40 & 75.96 \\
48 & 6 & 8 & \textbf{80.41} & \textbf{76.96} \\
48 & 8 & 6 & 79.84 & 76.24 \\
36 & 6 & 6 & 79.44 & 75.42 \\\hline   
\end{tabular}}    
\end{center}    
\end{table}

\begin{figure}[!t]
\centering
\includegraphics[width=1.0\columnwidth]{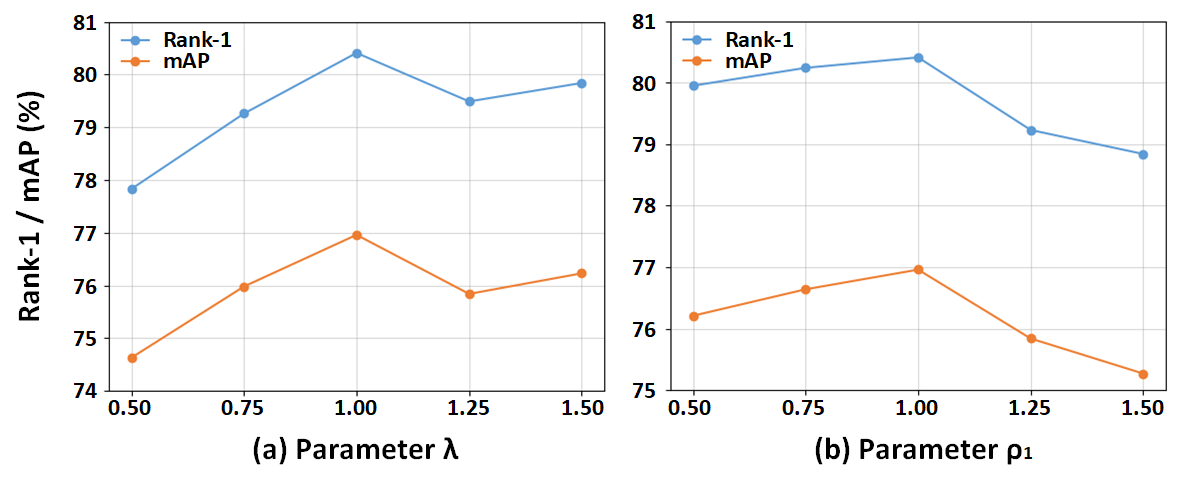}
\caption{Analyze the effect of varying $\lambda$ and $\rho_{1}$ values on the SYSU-MM01 dataset.}
\label{fig3}
\end{figure} 

\begin{figure*}
\centering
\includegraphics[width=1.9\columnwidth]{tsne}
\caption{(a)-(c) show the distance distribution between cross-modal features. Blue and green represent the distance frequency distribution of intra-class and inter-class respectively, and the vertical line denotes the distance mean of the distribution. (d)-(f) show the feature space distribution. Distinct colors and shapes signify various identities and modalities, while dashed ellipses indicate areas of identity confusion.}
\label{tsne}
\end{figure*}

\textbf{Influences of different shallow features, loss functions and combinations.} To identify the optimal MPDE configuration, we first utilize the shallow features output from the initial four stages as inputs of the module and then gradually remove the features output by the earliest stage. Meanwhile, we also investigate the model’s performance under different loss function constraints such as Triplet loss and ICG loss. Notably, the Softmax operation in the mask attention calculation is replaced with a Sigmoid activation function in setting Stage (4), enabling spatial feature-adaptive weighting. 

The experimental results presented in Table \uppercase\expandafter{\romannumeral6} (row 6) demonstrate that our method achieves the best performance when combining the Identity Clues Guided (ICG) loss with shallow features from the outputs of Stage-2, Stage-3 and Stage-4. Notably, incorporating features from all four stages (Stage (1,2,3,4)\textsuperscript{*}) proves suboptimal for model performance. This can be attributed to the adoption of independent weights in Stage-1 to handle heterogeneous modalities, which can capture rich identity features in specific modality. While this design effectively captures rich intra-modal identity features, it also introduces substantial cross-modal feature discrepancies that hinder the subsequent feature refinement processes. Furthermore, ablation studies reveal that removing Stage-2 and Stage-3 from the weight-sharing architecture results in certain performance degradation compared to the Stage (2,3,4)\textsuperscript{*} configuration. This suggests that features extracted from shared branches at different network depths are beneficial for enhancing the discriminative power of modality-invariant representations through complementary hierarchical information. 

In addition, from the results in Table VI (row 1-4), we also see that the performance of all feature configurations degrades when employing Triplet loss, which shows that the ICG loss constraint outperforms the Triplet loss under the same shallow feature combination. This is due to the ICG loss uses smoother feature centers as constraints, effectively eliminating the style differences between modalities.

\textbf{Influence of different configurations on SDCE.} To investigate the individual importance of $Att{_{cro}}$, $Att{_{self}}$, the joint interaction block and the channel modulation factor in SDCE, we record the performance under different configurations. As shown in Table \uppercase\expandafter{\romannumeral7}, when both transformer blocks are cascaded, higher performance can be achieved compared with using only $Att{_{cro}}$. $Att{_{cro}}$ is utilized to distill identity-guiding knowledge from $\tilde{f}$, while $Att{_{self}}$ leverages this identity-guiding knowledge to enhance the discriminative power of modality-invariant features, creating a mutually beneficial relationship between the two blocks. Additionally, replacing the MLP in the two attention blocks with $JIB$ results in a 1.3\% improvement in Rank-1 accuracy. This indicates that suppressing noise information within channels can effectively enhance the feature refinement capability of the module. Incorporating $\beta_1$ into $Att{_{cro}}$ results in a certain performance improvement, while adding $\beta_2$ to $Att{_{self}}$ leads to a 0.52\% decrease in Rank-1 accuracy. This is because the $\beta_1$ in $Att{_{cro}}$ will suppress channel identity-irrelevant information. After the features are processed through $Att{_{cro}}$, these noises will be effectively removed, while further adding $\beta_2$ may lead to the degradation of identity semantics.

\begin{figure}[!t]
\centering
\includegraphics[width=0.95\columnwidth]{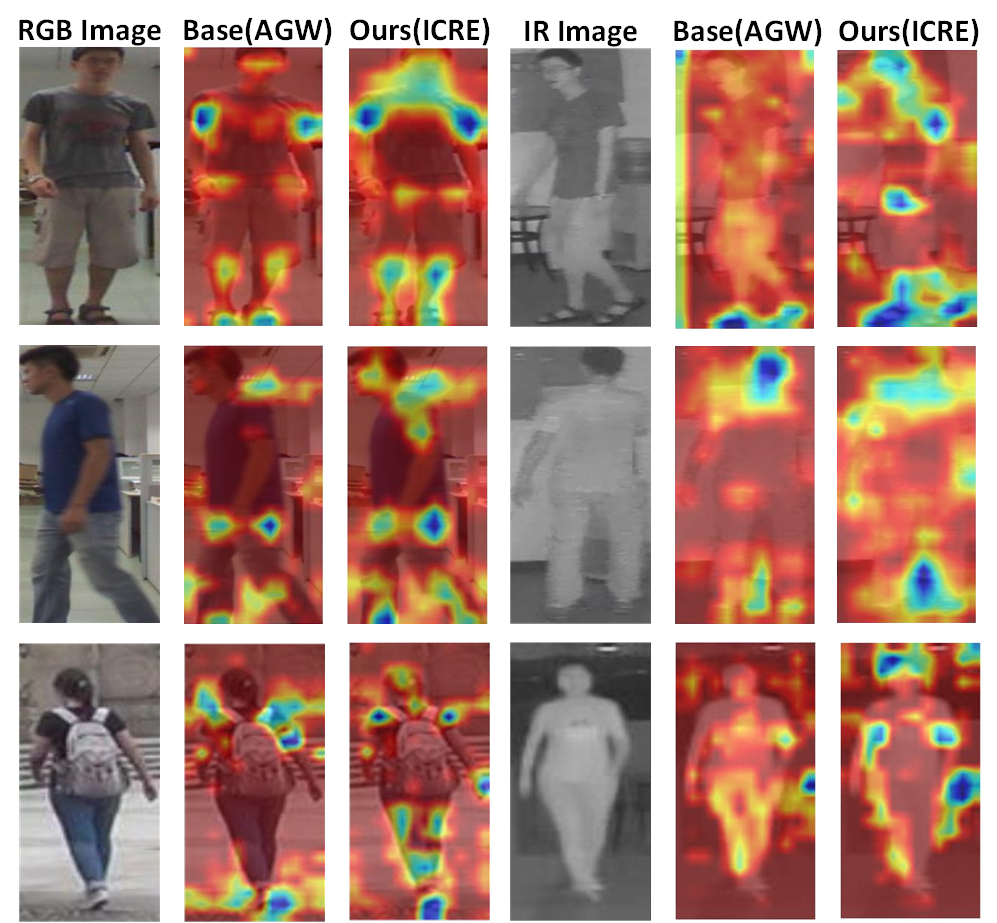}
\caption{The visualization outcomes of discriminative regions on sample images by Grad-CAM. Each row of visible and infrared images belongs to the same identity, and the baseline visualization results are added for comparison.}
\label{GradCAM}
\end{figure}

\begin{figure}[!t]
\centering
\includegraphics[width=0.95\columnwidth]{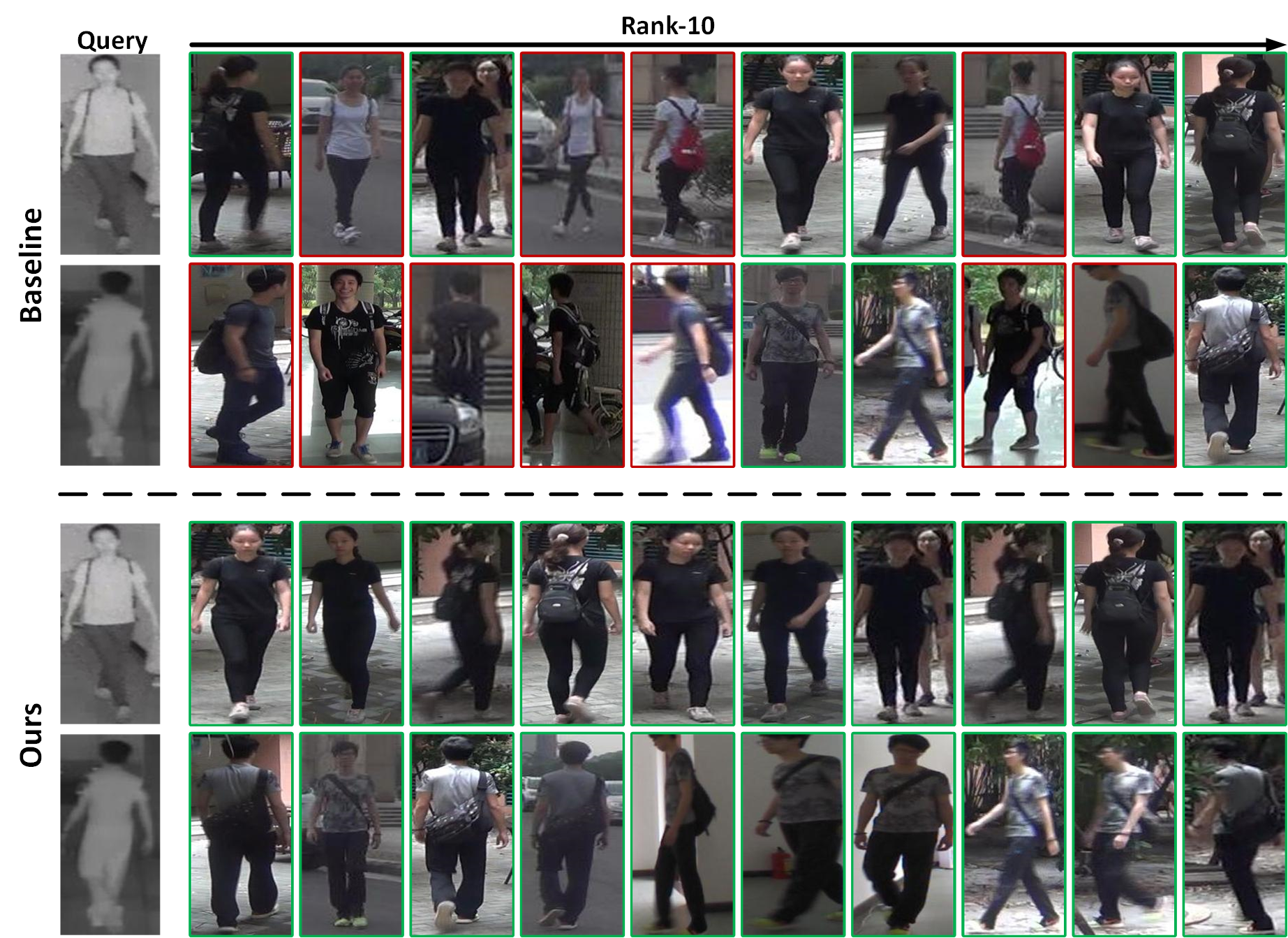}
\caption{Some search results of our ICRE and baseline on the SYSU-MM01 dataset. We retain the top-10 most similar samples, with green and red boxes representing true and false retrieval results respectively.}
\label{ranking}
\end{figure} 

\textbf{Analysis of hyperparameters $\bm{\lambda}$ and $\bm{{\rho }_{1}}$.} We provide a quantitative analysis and comparison in Figure~\ref{fig3} to evaluate the influence of various hyperparameter values. An appropriate ${{\rho }_{1}}$ can reduce the strength of inter-modal constraints, enhance the diversity of sample representations in the shared modality space, and improve the model's generalization ability to difficult samples. It is evident that setting the hyperparameters $\lambda$ and ${{\rho }_{1}}$ to 1 and 0.01 can obtain the optimal performance.

\textbf{Computational and model complexity analysis.} Table~\ref{tab8} presents the parameter count, FLOPs, inference time, training time and Rank-1 accuracy of DEEN \cite{r14}, HOS-Net \cite{r36}, our proposed ICRE and their common baseline CAJ \cite{r38}, to comprehensively compare these methods. Based on the baseline method, our method achieves excellent performance with a slight increase in computational cost and model complexity. Specifically, with FLOPs and parameters less than half of those in HOS-Net, we achieve a Rank-1 accuracy of 80.41\%. Additionally, in terms of training time, our method demonstrates faster convergence speed and significantly reduces the GPU computational resources required for training.

\textbf{Influence of batch size on model performance.} The experimental results presented in Table \uppercase\expandafter{\romannumeral9} demonstrate that our method achieves the best performance when configured with 6 distinct identities, each represented by 8 images. Deviation from this configuration, either increasing or decreasing the batch size, consistently results in performance degradation. This phenomenon can be attributed to two primary factors: larger batch sizes tend to introduce greater intra-identity appearance variations, while smaller batch sizes are insufficient capture the necessary diversity to maintain robust generalization.

\begin{figure}[!t]
\centering
\includegraphics[width=1.0\columnwidth]{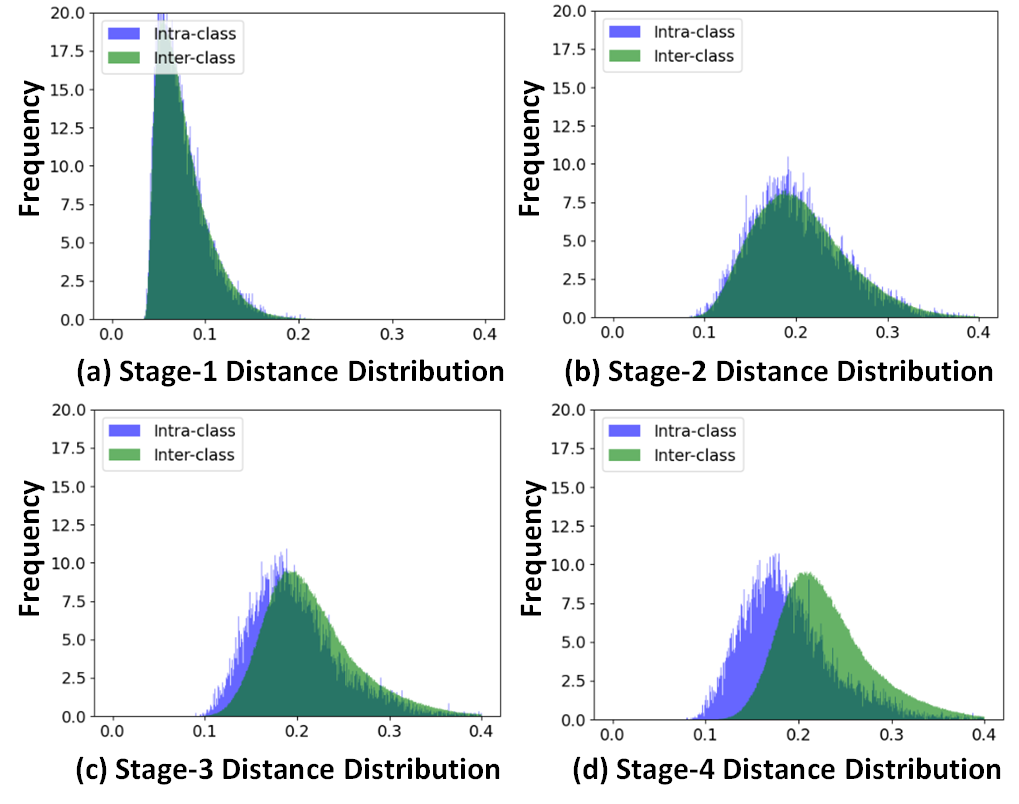}
\caption{Visualization of Shallow Feature Distance Distribution. (a)-(d) represent the distance distribution of output features of ResNet50 at different stages, respectively. Intra-class distances are depicted in blue, while inter-class distances are shown in green}
\label{fig7}
\end{figure}

\subsection{Visualization}
\textbf{Feature distribution.}
Feature distance serves as a metric for measuring the similarity and differences between samples. To showcase the identity discriminative ability of our model, we plot the frequency distribution of intra-class and inter-class distances, as shown in Figure~\ref{tsne} (a-c). Specifically, smaller intra-class distances indicate a stronger capability of the model to mitigate modality discrepancy, allowing it to correctly identify images of the same person across different modalities. Additionally, larger inter-class distances ($\delta$) suggest a lower likelihood of identity confusion. We find $\delta_{1}<\delta_{2}<\delta _{3}$, which means that gradually adding MPFR and SDCE on the basis of the baseline can significantly improve the network's ability to extract modality-invariant cues. We also use t-SNE to visualize the two-dimensional feature space to more intuitively verify the above statement. As shown in Figure~\ref{tsne} (d–f), the addition of MPFR and SDCE progressively reduces the overlap between different identity clusters, resulting in larger inter-class margins and more compact intra-class distributions.

\textbf{Model attention visualization.} 
To verify whether our proposed ICRE can focus more on identity clues guided by discriminative knowledge present in shallow features, we utilize Grad-CAM to generate attention heatmaps, which illustrate the areas of focus for the model. From Figure~\ref{GradCAM}, we can compare and conclude that our method focuses more on human-related areas and effectively removes background and other identity-irrelevant information, in contrast to the baseline. This is attributed to the fact that MPFR integrates modality-specific attributes and multi-scale semantics beneficial for identity discrimination from multiple shallow features. Additionally, SDCE effectively filters the integrated features and removes noise that could cause identity degradation.

\textbf{Retrieval results. }
Figure~\ref{ranking} presents some retrieval results from the testing process. We observe that compared to the baseline, our ICRE achieves higher retrieval accuracy. Thanks to the rich identity clues provided by modality-specific attributes and multi-scale semantics, ICRE can effectively mitigate the spectral diversity caused by different imaging processes. Even in the case of poor query image quality, it can accurately detect the same person based on details such as backpack straps or body structure.

\textbf{Distribution Visualization of Shallow Feature Distance.} To validate the effectiveness of aggregating shallow features in preserving modality-specific attributes, we visualize the intra-class and inter-class distance distributions of shallow features from different stages of ResNet50. As shown in Figure~\ref{fig7} (a), samples from different identities exhibit high distribution similarity in the shallow feature space. This is primarily due to the fact that modality-specific attributes introduce modality discrepancies, which hinder identity discrimination Further observations from Figure~\ref{fig7} (b)–(d) reveal that the overlap between inter-class and intra-class distributions gradually decreases with the increase of network depth. This indicates that deeper layers are capable of filtering modality-specific interference while focusing on shared discriminative semantics. This phenomenon confirms that shallow features inherently contain rich modality-specific attributes, providing theoretical support for our proposed ICRE module: 1) it captures modality attributes without requiring additional modality branches; 2) it achieves synergistic optimization between modality attributes and identity semantics through hierarchical feature fusion.

\section{Conclusion}
In this paper, we propose a novel Identity Clue Refinement and Enhancement (ICRE) network for VI-ReID, which captures identity-aware knowledge from shallow features and refines them into modality-invariant features. A Multi-Perception Feature Refinement (MPFR) module is designed to adaptively aggregate shallow features and it effectively captures modality-specific attributes and multi-scale semantics that are beneficial for identity discrimination. We further propose a Semantic Distillation Cascade Enhancement (SDCE) module to distill identity-aware knowledge from aggregated shallow features and guide modality-invariant feature learning. Additionally, our proposed Identity Clues Guided (ICG) Loss helps alleviate modality discrepancies and promotes diverse feature learning. Experimental results across multiple public datasets confirm the superiority of our ICRE approach over SOTA methods.

\bibliographystyle{IEEEtran}
\bibliography{ref}

\begin{IEEEbiography}[{\includegraphics[width=1in,height=1.25in,clip,keepaspectratio]{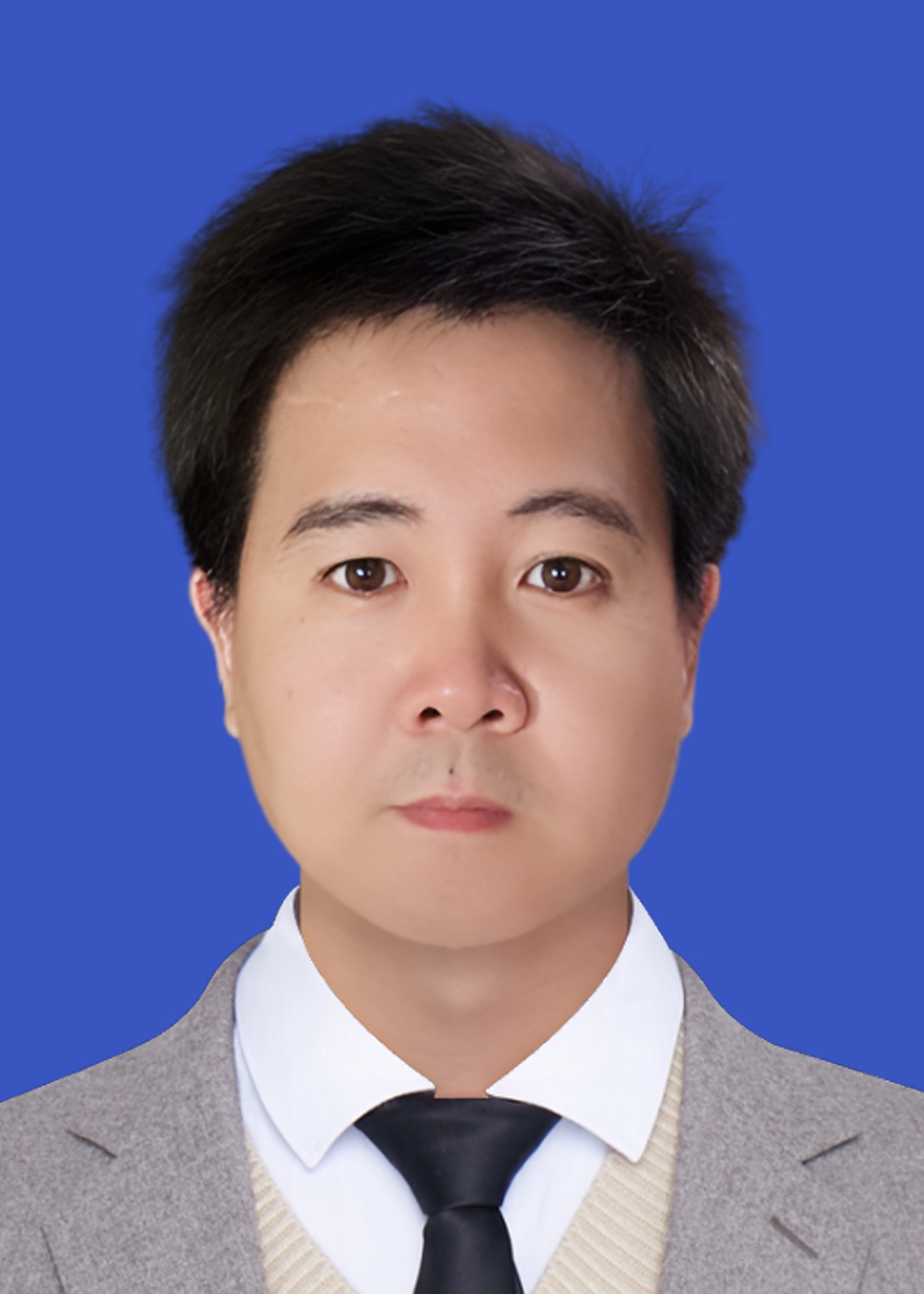}}]{Guoqing Zhang}
(Member, IEEE) received the B.S. and Master degrees in Information Engineering from the Yangzhou University, Yangzhou, China, in 2009 and 2012, and the Ph.D. degree in pattern recognition and intelligence system from Nanjing University of Science and Technology, Nanjing, China, in 2017. He is currently a Full Professor with the School of Computer science, Nanjing University of Information Science and Technology (NUIST), Nanjing, China. His current research interests include computer vision, pattern recognition and machine learning.
\end{IEEEbiography}

\begin{IEEEbiography}[{\includegraphics[width=1in,height=1.25in,clip,keepaspectratio]{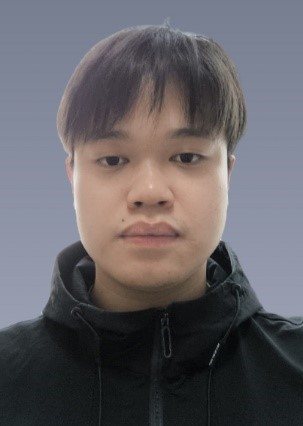}}]{Zhun Wang}
received the B.E. degree from Nanjing XiaoZhuang University, China, in 2022. He is currently pursuing a master's degree at the School of Software Engineering, Nanjing University of Information Science and Technology. His current research interests include computer vision, image processing, cross-modality learning and person re-identification.
\end{IEEEbiography}

\begin{IEEEbiography}[{\includegraphics[width=1in,height=1.25in,clip,keepaspectratio]{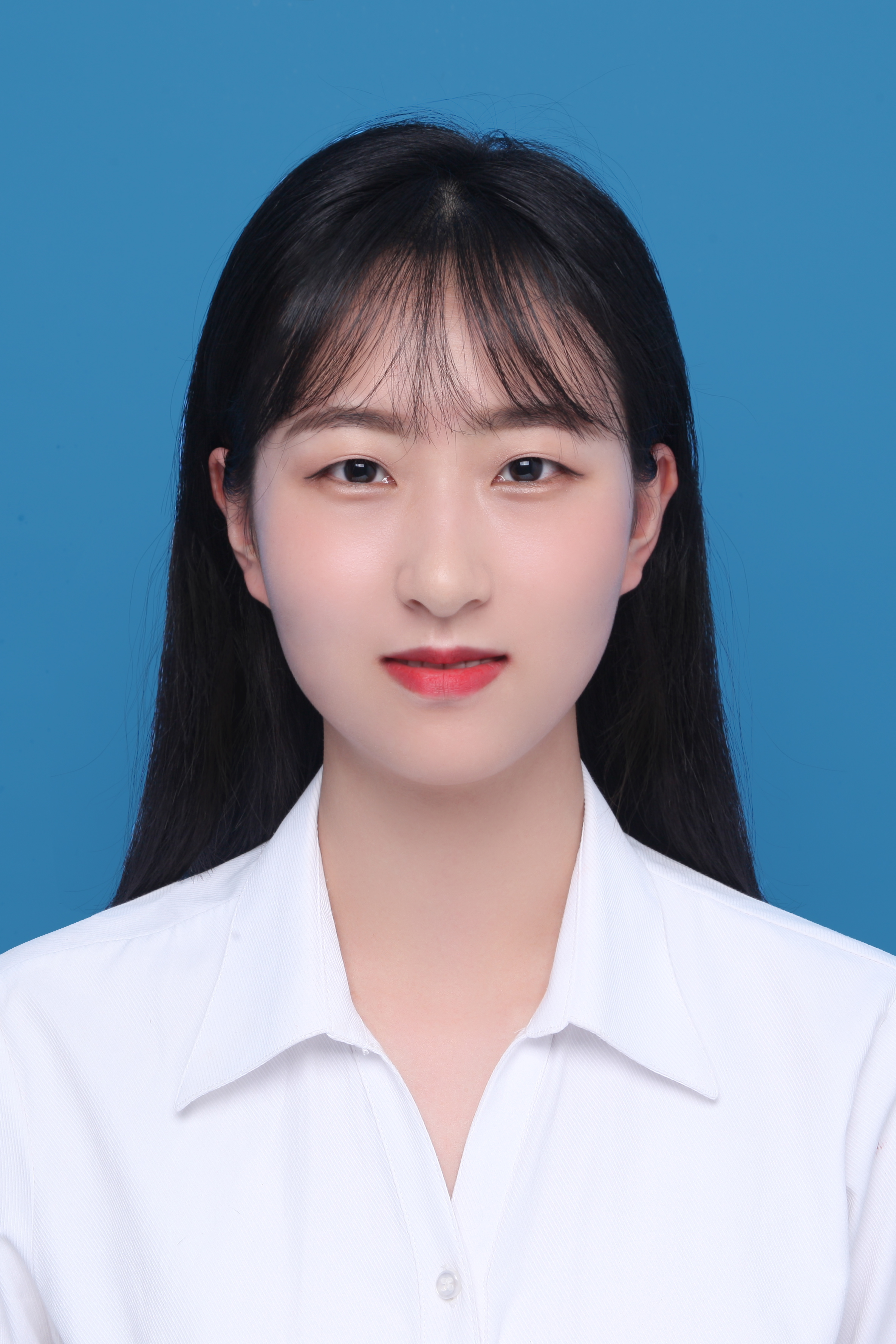}}]{Hairui Wang}
received the B.E. degree from the School of Computer Science, Nanjing University of Information Science and Technology, NanJing, China, in 2022, where she is studying for a master’s degree now. Her research interests include computer vision, Person Re-ID and related applications.
\end{IEEEbiography}

\begin{IEEEbiography}[{\includegraphics[width=1in,height=1.25in,clip,keepaspectratio]{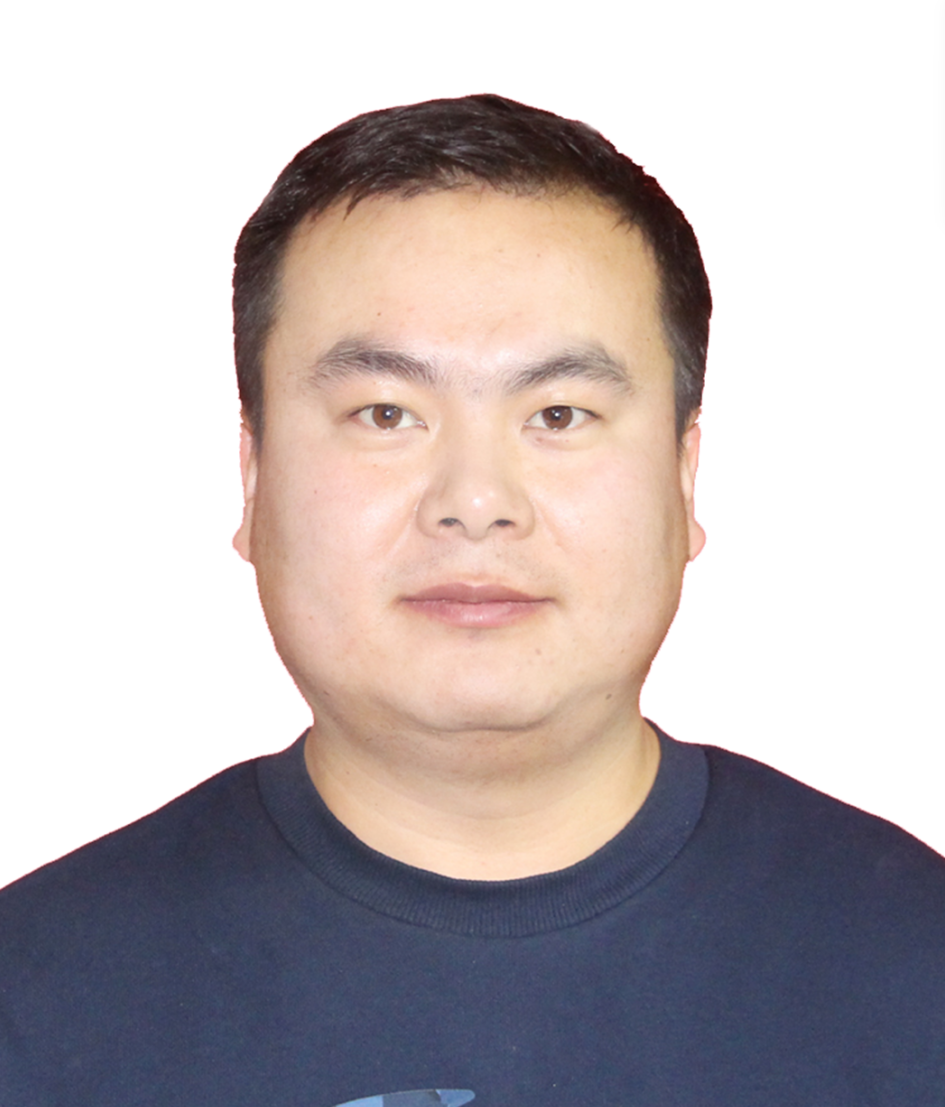}}]{Zhonglin Ye}
was born in 1989. He received the Ph.D. degree from Shaanxi Normal University, Xi’an, China, in 2019. He is currently a Post Doctoral Fellow with the Nagasaki Institute of Applied Science, Japan. He is also an Associate Professor with the School of Computer Science, Qinghai Normal University, Xining, China. He is also a member of the State Key Laboratory of Tibetan Intelligent Information Processing and Application. His main research interests include data mining, knowledge discovery, graph neural network, network representation learning, and machine learning.
\end{IEEEbiography}

\begin{IEEEbiography}[{\includegraphics[width=1in,height=1.25in,clip,keepaspectratio]{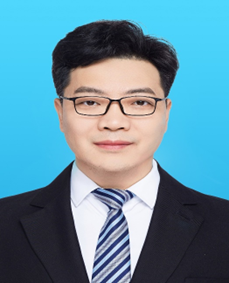}}]{Yuhui Zheng} (Member, IEEE) was born in Shanxi, China, in 1982. He received the B.Sc. degree in pharmacy engineering and the Ph.D. degree in pattern recognition and intelligent system from Nanjing University of Science and Technology, Nanjing, China, in 2004 and 2009, respectively. From 2014 to 2015, he was a visiting professor in the digital media laboratory of the school of Electronic and Electrical Engineering, Sungkyunkwan University, Korea. He is currently a Full Professor at the State Key Laboratory of Tibetan Intelligent Information Processing and Application, Qinghai Normal University. His main research areas include image and video analysis, scene understanding, visual tracking, and pattern recognition.
\end{IEEEbiography}

\end{document}